\documentclass[runningheads]{llncs}
\usepackage[T1]{fontenc}
\usepackage{graphicx}
\usepackage{amsmath}
\usepackage{epstopdf}
\usepackage{color}
\usepackage{comment}
\usepackage{url}
\usepackage{booktabs}
\usepackage{textgreek}
\usepackage{multirow}
\usepackage{float}
\usepackage{amsfonts}
\usepackage[table]{xcolor}
\usepackage{tikz}

\definecolor{myred}{rgb}{.8,.0,.0}

\newcommand{\MDD}{M100}
\newcommand{\MBD}{F25M75}
\newcommand{\ESD}{F50M50}
\newcommand{\FBD}{F75M25}
\newcommand{\FDD}{F100}

\begin{document}
%
\title{Dataset Distribution Impacts Model Fairness: Single vs. Multi-Task Learning}
%
%
\ifdefined\DOUBLEBLIND
    \author{Anonymised Authors\inst{1,2,3}\\
    *****
    **** \inst{}}
    \authorrunning{Anonymised Authors}
    \institute{*** \and
    *** \and
    *** }
\else
    \author{Ralf Raumanns\inst{1,2}\and
    Gerard Schouten\inst{1} \and
    Josien P. W. Pluim\inst{2} \and \\
    Veronika Cheplygina\inst{3}}
    \authorrunning{R. Raumanns et al.}
    \institute{Fontys University of Applied Science, Eindhoven, The Netherlands \and
    Eindhoven University of Technology, Eindhoven, The Netherlands \and
    IT University of Copenhagen, Denmark}
\fi

\maketitle              
\begin{abstract}
The influence of bias in datasets on the fairness of model predictions is a topic of ongoing research in various fields. We evaluate the performance of skin lesion classification using ResNet-based CNNs, focusing on patient sex variations in training data and three different learning strategies. We present a linear programming method for generating datasets with varying patient sex and class labels, taking into account the correlations between these variables. We evaluated the model performance using three different learning strategies: a single-task model, a reinforcing multi-task model, and an adversarial learning scheme. Our observations include: 1) sex-specific training data yields better results, 2) single-task models exhibit sex bias, 3) the reinforcement approach does not remove sex bias, 4) the adversarial model eliminates sex bias in cases involving only female patients, and 5) datasets that include male patients enhance model performance for the male subgroup, even when female patients are the majority. To generalise these findings, in future research, we will examine more demographic attributes, like age, and other possibly confounding factors, such as skin colour and artefacts in the skin lesions. We make all data and models available on GitHub.

\end{abstract}

\section{Introduction}\label{Introduction}



Deep learning has shown many successes in medical image diagnosis \cite{Saha2024-nq,Esteva2017-hn,Ehteshami_Bejnordi2017-gr}, but despite high overall performance, models can be biased against patients from different demographic groups \cite{abbasi2020risk,larrazabal2020gender,gichoya2022ai}. Bias and fairness are becoming an active topic in medical imaging, with studies focusing for instance on skin lesions \cite{abbasi2020risk,Groh2021-vm}, chest x-rays  \cite{larrazabal2020gender} and brain MR scans \cite{petersen2022feature}. Examples of sensitive attributes include age, sex or race. For skin lesion classification, the Fitzpatrick skin type is often studied \cite{Seth2024-ai,Bencevic2024-pg,Groh2021-vm,Wu2022-hl}.

Fairness studies typically include baselines showing bias between groups, and/or propose methods to improve fairness. The methods are based on sampling or weighting strategies during training \cite{Groh2021-vm}, and/or introducing training strategies that try to debias the methods to rely on the sensitive attributes, such as adversarial methods \cite{abbasi2020risk}. For instance, Yang and colleagues \cite{Yang2023-cw} developed an adversarial debiasing framework to reduce biases in hospital location and patient ethnicity. Similarly, Wu et al. introduced FairPrune \cite{Wu2022-hl}, a method for pruning parameters based on their significance to both privileged and unprivileged groups, focusing on sex and skin tone. Moreover, Bevan and Atapour-Abarghouei \cite{Bevan2023-kg} employed various strategies to limit bias in skin lesion images, specifically targeting discrepancies arising from medical instruments, surgical markings, and rulers. Popular datasets include ISIC skin lesion datasets \cite{Gutman2016-lf,Codella2017-rd,Codella2019-cn,Tschandl2018-sz,Combalia2019-jj,Veronica2021patient} and  Fitzpatrick-17K \cite{Groh2021-vm,Groh2022-ja}. Researchers either use already provided data splits for evaluation, or split the data ratios of patients with a specific demographic attribute, for example male vs female patients.

Our current study builds on two crucial insights from other topics in medical imaging: multi-task learning and shortcut learning \cite{Geirhos2020-te,Nauta2021-fe}. Firstly, some studies use demographic attributes within multi-task learning settings; for example,\cite{liu2014iterated}. Here the attributes are \emph{reinforcing} the diagnosis during optimization. This is at odds with the more recent adversarial strategies \cite{Adeli2021-da,abbasi2020risk} where models are encouraged to NOT predict the sensitive attribute. Secondly, there are correlations between demographics and demographic attributes and shortcut learning,  including, for example, imaging devices and surgical markers \cite{willemink2020preparing,jimenez2023detecting,gichoya2022ai,Bissoto2020-hn,Bevan2023-kg}. In such cases, simply splitting the data according to a specific attribute can create imbalance in terms of the other attributes, thus the observed (un)fairness could be due to the attributes that were not considered. 

\noindent \textbf{Our contributions} are as follows:

\begin{enumerate}
\item We propose using the linear programming (LP) approach for skin lesions retrieved from the ISIC archive \cite{Gutman2016-lf,Codella2017-rd,Codella2019-cn,Tschandl2018-sz,Combalia2019-jj,Veronica2021patient} via the gallery browser \cite{Isic-query}. This method gives more control over patient subset assignment. It adjusts the proportions of selected dataset attributes while keeping others constant.

\item We systematically study two strategies that handle the demographic variable in different ways: a reinforcing multi-task strategy \cite{Marques2021-ck,raumannsr-melba} and an adversarial strategy \cite{abbasi2020risk,Adeli2021-da,Chu2022multitask}

\item We evaluate our models using overall and subgroup Area Under the Curve (AUC) based on sex, and show that:
\begin{itemize}
    \item Models perform better for male subgroups in the male-only and lightly skewed male patient experiments. In the balanced dataset and lightly skewed female patient experiments, there is no significant difference between the subgroups. However, in the lightly skewed female patient scenario, the adversarial model performs better for male patients.
    \item Models trained exclusively on female patients exhibit a positive difference in performance for female patients.
    \item The base model reveals a significant sex bias, performing worse for female patients, except when trained exclusively on female patients.
    \item The reinforcement model has no significant effect on sex bias.
    \item The adversarial model significantly reduces sex bias in scenarios involving only female patients.
    \item Eliminating model bias is challenging, with significant performance gaps observed in datasets with skewed sex distributions.
\end{itemize}

\item We make all data and models available on \url{https://github.com/raumannsr/data-fairness-impact}.
\end{enumerate}

\section{Methods}\label{Methods}
\textbf{Construction of datasets}. We trained and validated our models on datasets with different female (F) to male (M) patient ratios, equal numbers of malignant and benign lesions, and equal number of patients below and above 60 (median age) for each sex. We refer to the datasets as \MDD\ (100\% male patients), \MBD\ (25\% female patients, 75\% male patients), \ESD, \FBD, and \FDD\ are defined analogously. We evaluate the models using a balanced test set mirroring \ESD. 

We used the ISIC archive's \cite{Gutman2016-lf,Codella2017-rd,Codella2019-cn,Tschandl2018-sz,Combalia2019-jj,Veronica2021patient} gallery browser \cite{Isic-query}, which had 81,155 dermoscopic images of skin lesions, some with age and sex metadata. We queried the archive for ``dermoscopic'' images diagnosed as ``benign'' or ``malignant'' for all ages and both sexes. This gave us 71,035 images (62,439 benign, 8,596 malignant), which we processed using the steps in Fig. \ref{fig:sequential_steps} (see Appendix for more details). 
\begin{figure}
\centering
\begin{tikzpicture}[node distance=1.1cm, every node/.style={fill=blue!5, rounded corners, draw, text centered, align=center}]
  \node (A) [minimum width=9cm] {\scriptsize 1. Remove all lesions with an undefined age attribute.};
  \node (B) [below of=A, minimum width=9cm] {\scriptsize 2. Remove duplicates following Cassidy's approach \cite{Cassidy2022-wx}.};
  \node (C) [below of=B, minimum width=9cm] {\scriptsize 3. Remove multiplets by selecting a random image from each patient.};
  \node (D) [below of=C, minimum width=9cm] {\scriptsize 4. Use linear programming to find the optimal sampling according\\
  \scriptsize to sex, age (above or below median), and class label.};
  \node (E) [below of=D, minimum width=9cm, fill=blue!5] {\scriptsize 5. Reserve 12.5\% (1264 lesions) of the balanced dataset as the test set.};
  \node (F) [below of=E, minimum width=9cm, fill=blue!5] {\scriptsize 6. Subsample the remaining data into subsets of \\
  \scriptsize 2206 malignant and 2206 benign instances each, see Table \ref{tab:result_lp_model}.};
  \node (G) [below of=F, minimum width=9cm, fill=blue!5] {\scriptsize 7. Use the subsets to create training and validation sets with\\
  \scriptsize 3528 and 884 lesions respectively.};

  \draw[->] (A) -- (B);
  \draw[->] (B) -- (C);
  \draw[->] (C) -- (D);
  \draw[->] (D) -- (E);
  \draw[->] (E) -- (F);
  \draw[->] (F) -- (G);
  \draw[->, rounded corners] (G.east) -- ++(1,0) -- ++(0,2.2) -- (E.east);
\end{tikzpicture}
\caption{Steps for filtering lesions and creating test, training and validation sets. Steps 5 through 7 are repeated using 5 different seeds in a cross-validation setup.} \label{fig:sequential_steps}
\end{figure}\\\\
\textbf{Linear programming for optimal dataset construction}. We have developed a method to create diverse dataset compositions using linear programming, a common mathematical optimisation technique. The goal is to maximise the number of instances of skin lesions within defined constraints, as we express below:

\begin{equation*}
\begin{array}{ll@{}ll}
\text{Find a vector} & & x & \text{(decision variables)}\\
\text{that maximises}  & & f=x_1 & \text{(objective\ function)}\\
\text{subject to} & & a_{i1}x_1 + a_{i2}x_2 + \dots + a_{in}x_n\leq b_i & \text{(constraints)}\\
& & \text{for } i=1,\dots,13 &\\
\text{and} & & x_j\geq 0. & \text{(non-negativity constraints)}\\
& & \text{for } j=1,\dots,14 &\\
\end{array}
\end{equation*}

\begin{itemize}
    \item The decision variables ($x_1,\dots,x_{14}$) denote specific categories, like benign lesions in female patients aged 60 and above. See Appendix for more.
    \item The objective function of the LP model is designed to maximise the number of malignant instances $x_1$. There are fewer malignant instances than benign ones in the ISIC archive, and the goal is to achieve a balance between the two.
    \item The constraints limit the solution by setting specific limits for each group and maintaining ratios between these groups. Group examples include all benign lesions, all female patients over 60, and all male patients under 60. The primary constraint ensures an equal number of malignant and benign lesions ($x_1 - x_2 = 0$). See Appendix for more.
    \item Due to non-negativity constraints, decision variables cannot be negative.
\end{itemize}

\begin{table}[hbt!]
\caption{Datasets are distributed amongst malignant, benign, male patients (M), and female patients (F) categories for both training and validation.}
\resizebox{\textwidth}{!}{  
\setlength{\tabcolsep}{3pt}
\begin{tabular}{llllll}
& \MDD & \MBD & \ESD & \FBD & \FDD \\
\hline
Malignant (M/F) & 2206 (2206/0) &  2941 (2206/735) & 4412 (2206/2206) & 3235 (809/2426) & 2426 (0/2426)\\
Benign (M/F) & 2206 (2206/0) & 2941 (2206/735) & 4412 (2206/2206) & 3235 (809/2426) & 2426 (0/2426)\\
\end{tabular}
}
\label{tab:result_lp_model}
\end{table}

Within set constraints, the optimal solution maximises malignant lesions and assigns value to decision variables. To find this solution, we created a unique LP model for each dataset. Table \ref{tab:result_lp_model} shows the result of the LP model for the different datasets.\\

\noindent\textbf{Models}. We used the ResNet50 model \cite{He2016-ow} in three ways, which include:
\begin{itemize}
    \item The single-task baseline model enhanced with two fully connected layers has a sigmoid activation function and binary cross-entropy loss ($L_{c}$,  see Equation \ref{eq:aeq_0}). We did not use class weights but rather the actual distribution represented by the training dataset. We fine-tuned the model through a grid search of three varying learning rates and batch sizes, selecting the combination that yielded the highest performance across all experiments.
    \item The multi-task reinforcing model, with three added layers to the convolutional base, produces two outputs: one for classification and the other for the binary sex attribute. We employed binary cross-entropy loss ($L_{c}$) and a sigmoid activation function for both heads, giving equal weight to both losses.
    \item  The multi-task adversarial model was implemented following the methodology \cite{Adeli2021-da,abbasi2020risk}, using a network with a shared feature encoder and two classifier heads. One classifier targeted skin cancer classification; the other predicted confounding variables like sex or age. We used the ResNet architecture to compare performance with baseline and reinforcing models equitably. We trained the skin cancer classifier and encoder with cross-entropy loss ($L_c$) and optimised the bias predictor with binary cross-entropy loss ($L_c$). To diminish the confounder predictiveness of the encoded feature, we adversarially adjusted the encoder using a third loss ($L_{br}$), setting λ as the penalty for accurate demographic predictions. We used a lambda (λ, see Equation \ref{eq:aeq_1}) value of 5, as in  \cite{abbasi2020risk}, to assess subgroup performance and set the penalty for correct predictions of the target demographic parameter.
\end{itemize}
To summarise, we use these loss functions:
\begin{equation}\label{eq:aeq_0}
L_{c} = -\sum_{i=1}^{n} [y_i \log(\hat{y_i}) + (1 - y_i) \log(1 - \hat{y_i})]
\end{equation}
\begin{equation}\label{eq:aeq_1}
L_{br}= \lambda{L_{c}} 
\end{equation}

where \(n\) represents the number of lesions, and \(y_i\) and \(\hat{y_i}\) denote respectively the prediction and the expected outcome for lesion \(i\).

We pre-trained all networks with ImageNet and resized images to 384 × 384 for ResNet50's input size. During model training, we used data augmentation techniques, ran up to 40 epochs with a batch size of 20, and set a learning rate of $2.0\mathrm{e}{-5}$. We stopped training if no significant improvement occurred after 10 consecutive epochs to avoid overfitting. We implemented our baseline and reinforcing models in Keras with the TensorFlow backend \cite{Geron2022-bm} and our adversarial model in PyTorch \cite{Paszke1912-ny}.\\\\
\textbf{Evaluation.} 
For the purpose of an in-depth evaluation, we generated five distinct instances of each dataset: \MDD, \MBD, \ESD, \FBD\ and \FDD. Each instance was created with a unique seed to ensure diversity and robustness in our evaluation. Each seed corresponds to a balanced test set to allow fair comparisons between dataset-model instances. Furthermore by using a balanced test set, we ensure that the results are not skewed towards any specific subgroup. We evaluate AUC overall and for male and female subgroups for each model architecture and dataset combination.

\begin{figure}[!ht]
    \centering
    \includegraphics[width=1\textwidth] {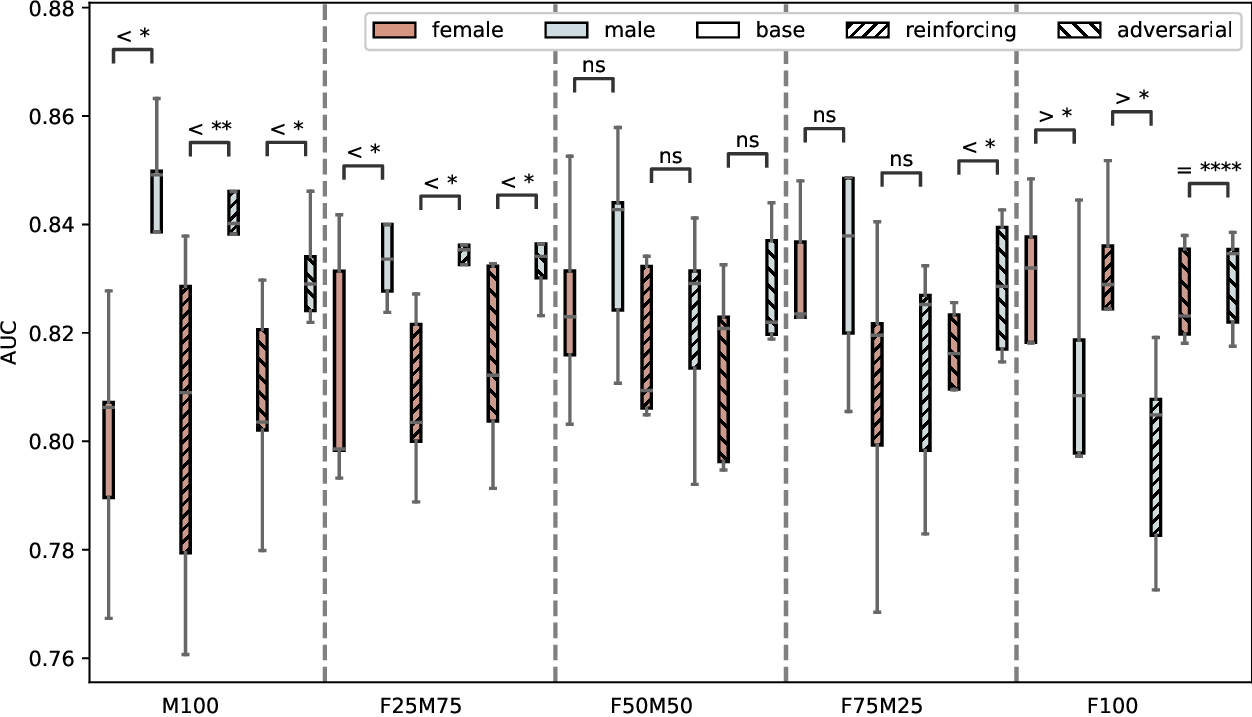}
    \caption{The AUC score varies based on data splits ranging from only male patients (\MDD) to only female patients (\FDD). We show base, reinforcing and adversarial model performance for female and male patient subgroups. Significance per Mann–Whitney U test (as used in \cite{larrazabal2020gender}) is denoted by **** \((P \leq 0.0001)\), *** \((0.0001 < P \leq 0.001)\), ** \((0.001 < P \leq 0.01)\), * \((0.01 < P \leq 0.1)\), and not significant (ns) \((P > 0.1)\). < indicates lower AUCs, > higher AUCs, and = comparable AUCs for female patients.}
    \label{fig:boxplot_reinforcing_adversarial_task}
\end{figure}

\section{Results}\label{Experimental results}
Figure \ref{fig:boxplot_reinforcing_adversarial_task} shows the impact of dataset distributions on three learning strategies, reporting AUC scores overall and for both sexes.\\

\noindent\textbf{Sex-specific training data yields better results.} Models perform better for male subgroups in the male-only and lightly skewed male patient experiments. In the balanced dataset and lightly skewed female patient experiments, there is no significant difference between the subgroups. However, in the lightly skewed female patient scenario, the adversarial model performs better for male patients. An exception is observed when the training datasets consist only of female patients; in such cases, there is a positive difference in AUC scores favouring female patients. Thus, our models seem more attuned to male patients in mixed-sex training sets, irrespective of the percentage of female patients. The best results are achieved when both sexes are trained exclusively on their respective data. Despite this, a male subgroup bias is apparent as the results for female patients are significantly worse than for male patients when trained exclusively on their data. \\

\noindent\textbf{Base model reveals sex bias.} The base model shows a significant sex bias in performance. When only male patients are involved the base model reveals a substantial performance gap between male and female patients. The results are significantly worse for female patients in male-skewed scenarios, except for the female-only experiments. Interestingly, the base model performs better for female patients than the adversarial and reinforcing models in the \FBD\ dataset scenario. It is worth noting that the performance gap between the subgroups is not as apparent when the dataset includes male and female patients, unlike in the experiment that only involved male patients.\\

\noindent\textbf{Reinforcement model has no significant effect on sex bias.} Training the model only on male patients increases AUC score variability and reduces performance differences between both sexes. The reinforcement model does not significantly affect sex bias.\\

\noindent\textbf{Adversarial model eliminates sex bias in cases involving only female patients.} The adversarial model reduces sex bias significantly in scenarios with only female patients but is less effective in other scenarios, often favouring male patients. Its performance varies across experiments and datasets.

\section{Discussion and conclusions}
We studied model and subgroup performance across datasets to identify the influence of bias in datasets on fairness of model predictions. We used linear programming (LP) to create various datasets with controlled male-female ratios. This was done to systematically evaluate the performance of three different learning strategies using ResNet-based CNNs. Other fairness and bias studies that require a flexible method to create datasets under certain constraints could benefit from this universal LP technique. 

Our study shows that eliminating bias is challenging. The adversarial model architecture is able to reduce sex bias in a female-only context but fails for other datasets. Other model approaches do not show convincing results with respect to bias reduction. 

Skewed sex distributions still show a performance gap between male and female patients. Our experiments demonstrate that the adversarial model better corrects sex bias in female-only datasets and not in male-only datasets, possibly due to other confounding and/or unidentified factors. Further research is needed on this issue.

As expected the base model shows a  sensitivity for sex bias, possibly due to overfitting. The reinforcing and adversarial models both having a form of regularisation (to counter overfitting), are potentially able to reduce sex bias compared to the base model. However in our experiments we only see a bias correction for adversarial models for female-only experiments.

Our outcomes show that sex-related information influences prediction tasks. Future research should determine which specific sex-related factors are essential to ensure fairness across different subgroups.

In contrast to categorical data like patient sex, where the groups are clearly defined, this is not possible or only partially possible with continuous data like age, which could lead to somewhat arbitrary subgroups. Therefore, we started with the demographic attribute sex and will continue similar research with the non-categorical age attribute.

Further, we have identified the following directions for future work:
\begin{itemize}
    \item Investigate whether using ``early stopping'' per task in a multi-task model reduces subgroup bias.
    \item Explore the impact of integrating segmentation with a classifier on sex-based disparities in identifying skin lesions.
    \item Study the roles of factors like skin colour and image artefacts in model fairness for different subgroups.
    \item Investigate the impact of shortcut learning on model fairness.
\end{itemize}

In conclusion, while we progress towards fairness, further advancements are needed to ensure consistent and equitable performance across various data distributions.

\begin{credits}

\ifdefined\DOUBLEBLIND
    \subsubsection{\ackname} [Blinded for review]
\else
    \subsubsection{\ackname} We gratefully acknowledge financial support from the Netherlands Organization for Scientific Research (NWO), grant no. 023.014.010.
\fi

\subsubsection{\discintname}
The authors declare that they have no known competing financial interests or personal relationships that could have influenced the work reported in this paper.

\end{credits}
%
%
%
%
\bibliographystyle{splncs04}
\bibliography{bibliography/refs_ralf,bibliography/refs_gerard,bibliography/refs_veronika}

\newpage
\noindent\textbf{Upper Boundaries LP Model}
\begin{table}
    \caption{The distribution of instances after the final step, which involves removing multiplets, establishes the upper boundaries of the decision variables in the LP model.}
    \centering
    \begin{tabular}{lrrr}
         & Aged & Female subgroup & Male subgroup \\
         \hline
         Malignant & $<60$ & 1,371 & 1,261\\
         & $\geq60$ & 1,641 & 2,801\\
         \hline
         Benign & $<60$ & 10,810 & 12,239 \\
         & $\geq60$ & 2,397 & 3,364\\
    \end{tabular}
    \label{tab:upper_boundaries_lp}
\end{table}

\noindent\textbf{Decision variables LP Model}\\\\
$x_{1}$:\text{\# malignant instances}\\
$x_{2}$:\text{\#  benign instances}\\
$x_{3}$:\text{\#  male patients (M) with malignant lesions}\\
$x_{4}$:\text{\#  female patients (F) with malignant lesions}\\
$x_{5}$:\text{\#  benign M }\\
$x_{6}$:\text{\#  benign F}\\
$x_{7}$:\text{\#  malignant lesions of M (age $<60$)}\\
$x_{8}$:\text{\#  malignant lesions of M (age $\geq 60$)}\\
$x_{9}$:\text{\#  malignant lesions of F (age $<60$)}\\
$x_{10}$:\text{\#  malignant lesions of F (age $\geq 60$)}\\
$x_{11}$:\text{\#  benign M (age $<60$)}\\
$x_{12}$:\text{\#  benign M (age $\geq 60$)}\\
$x_{13}$:\text{\#  benign F (age $<60$)}\\
$x_{14}$:\text{\# benign F (age $\geq 60$)}\\

\newpage
\noindent\textbf{Constraints LP Model}\\\\
\setcounter{equation}{0}
\begin{equation}
    x_{1}-x_{2}=0 \label{eq:aeq1}
\end{equation}
\begin{equation}
    rx_{4}-x_{3}=0 \label{eq:aeq2}
\end{equation}
\begin{equation}
    sx_{8}-x_{7}=0 \label{eq:aeq3}
\end{equation}
\begin{equation}
    tx_{10}-x_{9}=0 \label{eq:aeq4}
\end{equation}
\begin{equation}
    x_{1}-x_{3}-x_{4}=0 \label{eq:aeq5}
\end{equation}
\begin{equation}
    x_{3}-x_{7}-x_{8}=0 \label{eq:aeq6}
\end{equation}
\begin{equation}
    x_{4}-x_{9}-x_{10}=0 \label{eq:aeq7}
\end{equation}
\begin{equation}
    ux_{12}-x_{11}=0 \label{eq:aeq8}
\end{equation}
\begin{equation}
    vx_{14}-x_{13}=0 \label{eq:aeq09}
\end{equation}
\begin{equation}
    x_{2}-x_{5}-x_{6}=0 \label{eq:aeq10}
\end{equation}
\begin{equation}
    x_{6}-x_{13}-x_{14}=0 \label{eq:aeq11}
\end{equation}
\begin{equation}
    x_{5}-x_{11}-x_{12}=0 \label{eq:aeq12}
\end{equation}
\begin{equation}
    wx_{6}-x_{5}=0 \label{eq:aeq13}
\end{equation}
\\
\textit{Eq. \ref{eq:aeq1}}: \# of malignant records equals \# benign records.\\
\textit{Eq. \ref{eq:aeq2}}: The ratio $r$ of malignant male patients (M) to female patients (F).\\
\textit{Eq. \ref{eq:aeq3}}: The ratio $s$ of malignant M (age $<60$) to M (age $\geq 60$).\\
\textit{Eq. \ref{eq:aeq4}}: The ratio $t$ of malignant F (age $<60$) to F (age $\geq 60$).\\
\textit{Eq. \ref{eq:aeq5}}: \# of malignant lesions equals the \# M and F with malignant lesions.\\
\textit{Eq. \ref{eq:aeq6}}: \# M with malignant lesions is equal to that of all ages.\\
\textit{Eq. \ref{eq:aeq7}}: \# F with malignant lesions is equal to that of all ages.\\
\textit{Eq. \ref{eq:aeq8}}: The ratio $u$ of benign M (age $<60$) to M (age $\geq 60$).\\
\textit{Eq. \ref{eq:aeq09}}: The ratio $v$ of benign F (age $<60$) to F (age $\geq 60$).\\
\textit{Eq. \ref{eq:aeq10}}: \# benign lesions equals the \# M and F with benign lesions.\\
\textit{Eq. \ref{eq:aeq11}}: \# F with benign lesions is equal to \# all ages.\\
\textit{Eq. \ref{eq:aeq12}}: \# M with benign lesions is equal to \# all ages.\\
\textit{Eq. \ref{eq:aeq13}}: The ratio $w$ of benign M to F.

\end{document}